\documentclass[twoside,11pt]{article}

\usepackage{dnd}
\usepackage{times}

\usepackage{amsfonts}
\usepackage{amsmath}
\DeclareMathOperator*{\argmin}{arg\,min}
\usepackage{booktabs}
\usepackage{afterpage}
\usepackage{siunitx}
\usepackage{graphicx}
\usepackage{url}
\usepackage{hyperref}
\usepackage{longtable}
\usepackage{caption}
\dndheading{issue(number)}{2023}{firstpage--lastpage}{Nicholas Thomas Walker, Stefan Ultes, and Pierre Lison}{10.5087/dad.DOINUMBER}

\ShortHeadings{A Graph-to-Text Approach to Knowledge-Grounded Response Generation in HRI}{Walker, Ultes and Lison}
\firstpageno{1}

\begin{document}

\title{A Graph-to-Text Approach to Knowledge-Grounded Response Generation in Human--Robot Interaction}

\author{\name Nicholas Thomas Walker \email walker@nr.no \\
       \addr Norwegian Computing Center
       \AND
       \name Stefan Ultes \email stefan.ultes@uni-bamberg.de \\
       \addr University of Bamberg
       \AND 
       \name Pierre Lison  \email plison@nr.no\\
       \addr Norwegian Computing Center}

\editor{Name Surname}
\submitted{11/2023}{MM/YYYY}{MM/YYYY}

\maketitle

\begin{abstract}

Knowledge graphs are often used to represent structured information in a flexible and efficient manner, but their use in situated dialogue remains under-explored. This paper presents a novel conversational model for human--robot interaction that rests upon a graph-based representation of the dialogue state. The knowledge graph representing the dialogue state is continuously updated with new observations from the robot sensors, including linguistic, situated and multimodal inputs, and is further enriched by other modules, in particular for spatial understanding. The neural conversational model employed to respond to user utterances relies on a simple but effective \emph{graph-to-text} mechanism that traverses the dialogue state graph and converts the traversals into a natural language form. This conversion of the state graph into text is performed using a set of parameterized functions, and the values for those parameters are optimized based on a small set of Wizard-of-Oz interactions. After this conversion, the text representation of the dialogue state graph is included as part of the prompt of a large language model used to decode the agent response. The proposed approach is empirically evaluated through a user study with a humanoid robot that acts as conversation partner to evaluate the impact of the graph-to-text mechanism on the response generation. After moving a robot along a tour of an indoor environment, participants interacted with the robot using spoken dialogue and evaluated how well the robot was able to answer questions about what the robot observed during the tour. User scores show an improvement in the perceived factuality of the robot responses when the graph-to-text approach is employed compared to a baseline using inputs structured as semantic triples.

\end{abstract}

\begin{keywords}
Dialogue Management, Human-Robot Interaction, Large Language Models, Graphs, Task-oriented dialogue
\end{keywords}

\section{Introduction}

Recent advances in NLP have made large strides in improving the conversational abilities of dialogue systems \citep{bommasani2021opportunities}. Large language models (LLMs) based on the transformer architecture \citep{vaswani2017attention} have become the foundation for a large range of conversational agents, both task-oriented \citep{hosseini2020simple,peng2020few,nekvinda2022aargh} and open-domain \citep{zhang2020dialogpt,roller2021recipes,thoppilan2022lamda}. These advances have also raised questions about the capability of these models to perform tasks and reason about information \citep{valmeekam2022large}. Numerous investigations into the reasoning capabilities of LLMs have resulted in evidence of their abilities in some areas \citep{bubeck2023sparks}, while indicating shortcomings in others \citep{shi2023large, ullman2023large}. 

Meanwhile, there is a continuing interest and ongoing work in human-robot interaction (HRI) using large language models. A central question in this vein of research is how information about a robot's environment from different modalities can be effectively combined with LLMs. For instance, \cite{huang2022inner} demonstrated a system where feedback from the environment is used in the form of a kind of inner monologue. Under such a system, multiple sources of information are combined in a form representing the \emph{semantics} of a situation, in the sense of how objects under discussion relate to each other. These systems thereby make use of multiple modalities obtained from other (particularly pre-trained) models where the combined information enables improved performance even in zero-shot scenarios \citep{zeng2022socratic}. Such models suggest that there exists a strong capability for LLMs to operate over diverse sources of information structured as text input. 

In this article, we explore the problem of response generation for embodied agents and demonstrate how various sources of information can be collected and acted upon in such a system. Our approach rests on the following elements: 
\begin{enumerate}
    \item A dynamic knowledge graph of the dialogue state where locations, physical objects, and dialogue elements such as utterances are represented as nodes and semantic relations between them represented with edges.
    \item A simple parameterized function to verbalize the dialogue state graph based on traversals of the graphs, and the optimization of those verbalization parameters from Wizard-of-oz data.
    \item A large language model that includes the verbalized representation of the graph as part of its prompt and is responsible for generating the robot response. 
\end{enumerate}

\begin{figure}[t!]
    \centering
    \includegraphics[width=.95\textwidth]{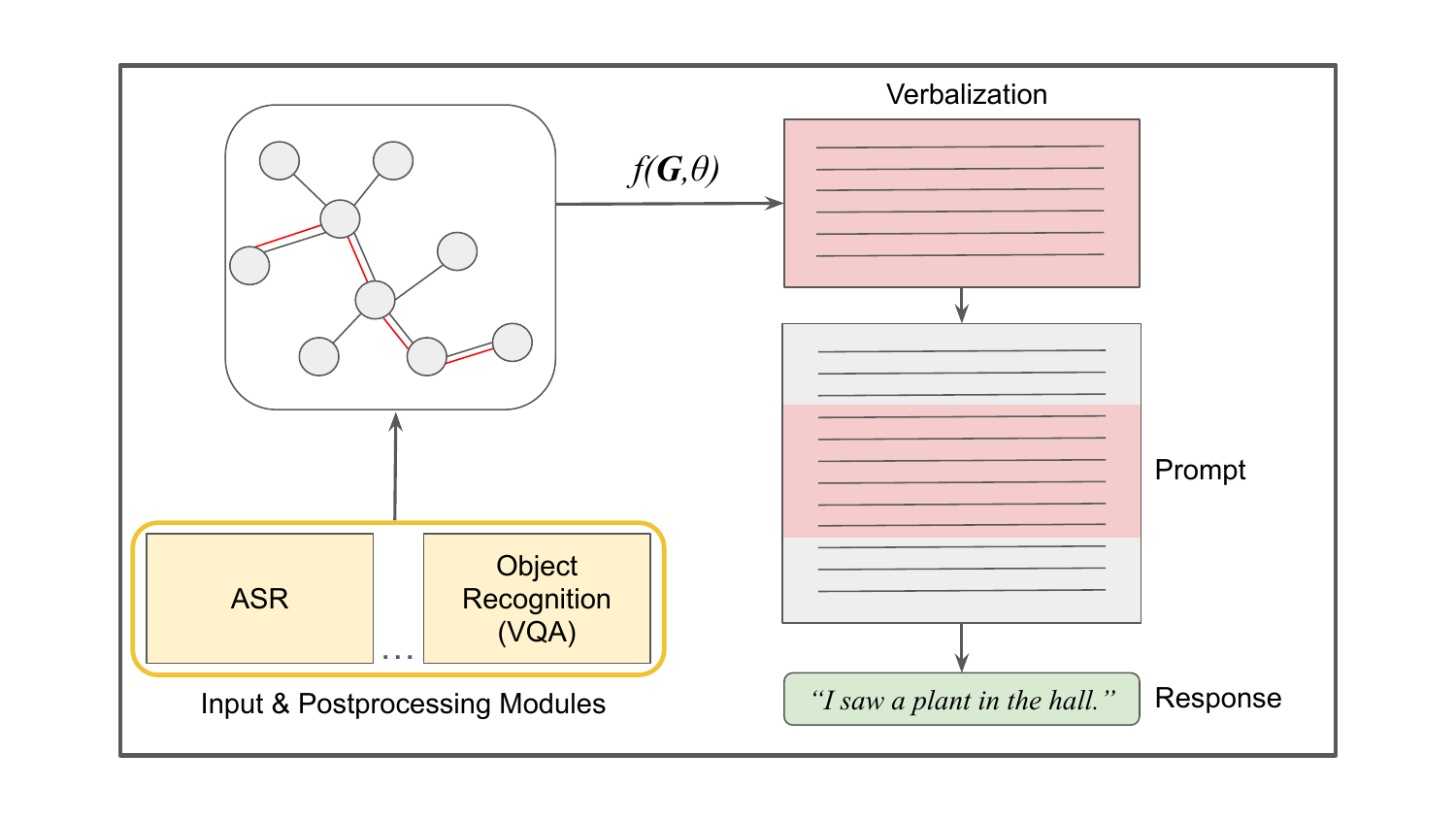}\vspace{-5mm}
    \caption{Diagram of the proposed approach. The red path in the graph indicates a traversal which is transformed to a natural language document via a parameterized function. The resulting text description is then inserted into the prompt for the language model.}
    \label{fig:approach}
\end{figure}

A sketch of the proposed approach is provided in Figure \ref{fig:approach}. As shown in the figure, information detected by the embodied agent including speech, visual, and positional data are brought together into a unified structure in the form of a graph and subsequently transformed into natural language as input for an LLM. Specifically, this is a \textit{dynamic} knowledge graph, in which nodes and edges are added incrementally. A dynamic rather than static graph structure allows it to be modified throughout robot exploration and conversation with users. The graph structure thereby enables the continuous inclusion of both situated and dialogue-specific information as it is received by the system. Semantic relations between objects, places, and utterances are represented as edges between the nodes of the graph, enabling not only verbal expression of these relations but also of larger \textit{traversals} of the graph. Moreover, probabilities and temporal relations in the graph are also easily exploited in such a transformation. The graph structure is thereby exposed to an LLM in a way that enables the strengths of such models to be exploited through a simple, controllable transformation from graph to text. 

To evaluate the proposed approach, we design an experiment with users to explore its use in a practical setting. Our experiment consists of users touring an office space with a robot and subsequently discussing the tour. We evaluate the robot's ability to converse about the tour when equipped with two LLMs (specifically GPT-4 and LLaMA), using the graph-to-text approach and a baseline triples conversion. From the resulting dialogues and user feedback, we provide a statistical analysis of the results along with further qualitative analysis of attributes of the resulting dialogues.

The rest of the paper is structured as follows. In Section \ref{sec:background}, we provide a general background on embodied conversational agents, knowledge graphs for dialogue, and large language models. Section \ref{sec:approach} describes how to structure the dialogue state of the agent as a knowledge graph, verbalize its content through graph traversals, and integrate the result into the prompt of the LLM employed to generate the robot responses to user utterances. Section \ref{sec:experiments} presents an experimental study that investigates the influence of this graph-to-text approach on the factuality and adequacy of the robot responses. The obtained user ratings show that this graph-to-text approach is able to improve the factuality of the robot responses compared to a baseline relying on inputs where the graph information is provided in the form of semantic triples. Finally, in sections \ref{sec:discussion} and \ref{sec:conclusion} we discuss the experimental results and review the benefits and limitations of the proposed approach, and conclude. 
 
\section{Background}
\label{sec:background}

In this section, we discuss the background to our study, starting with contemporary approaches to embodied agents in HRI. We then discuss recent work where graphs have been used with dialogue systems to represent situated dialogue information. Finally, we review the recent usage of LLMs in dialogue systems and the variety of approaches to presenting background information to them, including with graphs and for embodied agents. 

\subsection{Embodied Conversational Agents}
\label{sec:embodied_agents}

Embodiment refers broadly to the operation of an artificial agent in physical space, though specific definitions vary in precision and strictness \citep{miller2016embodiment}. Whichever definition may be preferred, embodiment has a substantial impact on dialogue and affects how speakers perceive and relate to the dialogue \citep{wainer2006role, pickering2009prediction}. In human--robot interaction, grounding the dialogue in the physical environment is often a prerequisite to language understanding. Grounded language, in the sense employed by e.g.~\citet{tellex2020robots}, refers to situated language, i.e.\ language with meaning in a physical environment.


In a situated scenario, dialogue between a human and a dialogue agent has a direct spatial relation to the surrounding environment. An embodied, robotic agent must take into account numerous input sources from the environment as well as any human users to accomplish complex tasks requiring both natural language or physical outputs, i.e. movement. Beyond spatial data and language input, \emph{visual} perceptions of varying kinds are an important source of information for the dialogue. 

These kinds of information may also be viewed in relation to time. For instance, the utterances in a dialogue are naturally temporally related. Moreover, embodied agents may take different actions depending on what information is most recently observed. Linguistic, visual, and spatial information acquired by the robot may also be \emph{uncertain}, presenting an additional challenge in effectively leveraging world knowledge to accomplish tasks. The agent's grounding in the physical world is subject to uncertainty arising from both perceptual module errors as well as inherently uncertain aspects of the environment. Thus, temporality and probability interact with these kinds of data to present a nuanced picture of the world for the agent to act upon. When quantified and represented by the system, uncertainty about the environment or dialogue state can be communicated to the user \citep{hough2017s}. By dialogue state, we refer to a representation of the current dialogue context which contains all information that is relevant for responding to user utterances.

Within HRI, language input is one communication medium among many that can be used to interact with a robot \citep{goodrich2008human}. In some cases, it may be used to instruct a robot to perform actions in its physical environment. For example, \cite{rankin2021robotic} demonstrated semantic parsing of instructions to associate the instructions with desired locations for the robot to explore. In a smaller setting, \cite{lynch2023interactive} explored the use of imitation learning through natural language instructions to guide a robotic arm to move objects. The real-time task required an approach that could account for both spatial information and the temporal ordering of actions and instructions given to the robot. Another model integrating spatio-temporal grounding was presented by \cite{shah2023lm}. In their model, a robotic navigation system combines a pre-trained visual navigation module with an LLM to construct a path following instructions. This approach first identifies landmarks with an LLM and subsequently grounds this information in a set of visual inputs from the navigation area using a second module. Each of these approaches required alignment of language input to the robot's physical environment. 




\subsection{Knowledge graphs for dialogue}

\subsubsection{Graph-based representations}
  
While dialogue state information may be represented in numerous ways, graphs are a flexible data structure with desirable properties with regard to both natural language and the physical world. A graph $\textit{G}$ consists of a set of vertices (or nodes) $\textit{V}$ paired with a set of edges $\textit{E}$. Each edge describes a pair of vertices, which may be ordered or unordered (describing a directed or undirected graph, respectively). Knowledge graphs specifically represent entities, that is to say objects and concepts, as well as relations between them. An example of graph-structured background information for dialogue is the dataset of \cite{eric-etal-2017-key}. In this dataset, each dialogue is associated with a knowledge base of entities for the dialogue system to retrieve in response to user queries. 
The dataset comprises three domains, of which the navigation domain represents entities with a spatial relation (distance) to the speaker and dialogue agent. With this property, the dialogue system can reason over the dialogue context to infer closest locations to recommend to the user. 
Another graph-based model of spatial information for embodied agents is the Scene Graph \citep{xu2020survey}, which represents physical objects as graph nodes and spatial relations as edges between them. Similarly, in the model presented by \cite{li2022embodied}, an embodied scene graph can also be generated by a model trained with imitation and reinforcement learning. The resulting graphs can then be applied to downstream tasks requiring spatial information. 

Within language data specifically, both utterance and dialogue level structures can be expressively rendered as graphs such as those of the Abstract Meaning Representation paradigm \citep{banarescu2013abstract}. A relevant use of this paradigm was presented by \cite{bonial2020dialogue}, extending the AMR schema for dialogue with a robot performing tasks in search and navigation. Their model of situated dialogue constructs graph representations of both user commands and agent responses. These graphs represent a semantic model of both objects in the environment and actions for the system to take. Graphs have also been used to represent dialogue structure as a whole and flows between utterances \citep{andreas2020task, gritta2021conversation}. Also specifically in HRI, \cite{wilcock2022conversational} presented a model for robotic conversational agents that used knowledge graphs to present semantic metadata to the system. 
Thus, in the context of previous work it is apparent that combining abstract and dialogue-specific information with situated knowledge provides a fruitful avenue for state representation \citep{papangelis2020towards}. Within this line of inquiry, our work builds upon a graph-structured dialogue state model wherein both dialogue elements and physical objects in the world are equally represented as nodes of a graph with edges in the graph representing semantic relations between them  \citep{walker2022management}.

Aside from the general usage of graphs representing linguistic structure or purely physical information, an additional kind of dialogue information that can be expressed in a graph is \emph{temporal} relations between objects. For instance, \cite{xing2023relational} introduced the DARER model, which used two types of temporal graphs representing semantic relations for dialogue act and sentiment prediction. In their model, user utterances are represented as nodes of a graph with ``previous" and ``future" relations between them. 
As demonstrated in a number of models \citep{tuan-etal-2019-dykgchat, he2021learning}, the dialogue agent's knowledge state can be captured as a dynamic knowledge graph that is updated throughout the dialogue. Dynamic, evolving graphs capture the intuition that the dialogue agent's knowledge should evolve along with the dialogue, encompassing perception of natural language utterances or physical observations as updates to the graph.

A further advantageous property of graphs is the capability of associating probabilities with the constituent nodes or edges of a graph. A common use of probability with graph structures in dialogue systems has been in handling uncertainty in action selection \citep{young2013pomdp}. Bayesian Networks are an additional example of a probabilistic model which has been used for dialogue management, including with multimodal data \citep{prodanov2003bayesian, thomson2008bayesian}. More generally, the problem of epistemic uncertainty \citep{celemin2023knowledge} about the dialogue state may also be usefully described by probabilities associated with graph elements. 

\subsubsection{Graph processing}

Although the flexibility of graphs allows for a wide diversity of state representations, it also requires adaptation of response generation models to operate on a graph structure. One means of doing so is the use of Graph Neural Networks (GNNs). An example of a GNN model used to process a large knowledge base was presented by \cite{christmann2023explainable}. Their model for conversational question answering used a GNN to construct a reduced representation of a large heterogeneous graph of entities. This model had the particular advantage of enabling explainability in the model along with using numerous combined sources of information. Particularly large knowledge graphs may require a filtering method or preprocessing to extract the most salient information \citep{wang2021improving}. GNNs have also been used among other things to \emph{verbalize} graphs, such as the method presented by \cite{yang2020program}, wherein a graph attention network is used with a verbalization of a program to verify facts. While GNNs have become an area of interest for using graph data in many domains, these models often require large quantities of in-domain training data and may require certain properties in the graph structure (e.g. homogeneity, connectivity). For instance, models such as the influential GraphSAGE of \cite{hamilton2017inductive} have limitations in their ability to distinguish certain non-isomorphic graphs \citep{garg2020generalization}. Moreover, GNNs also face difficulties when aggregating information across long paths in graphs, an ``over-squashing" of graph information explored by \cite{alon2020bottleneck}.

Graph verbalization is not limited to GNN-based models. \cite{xu2018sql} demonstrated a method of transforming Structured Query Language (SQL), itself viewable as a graph-structure, to natural language via a graph-to-sequence model. A common thread among approaches to verbalizing graphs is a correspondence of graph structures to natural language surface forms. A basic example is the conversion of a semantic triple (subject, predicate, and object) of two nodes and an edge between them to a sentence:
\begin{quote}
    \textsf{(`robot' $|$ `in' $|$ `hallway')}  $\rightarrow$ \emph{The robot is located in the hallway.}
\end{quote}
Such conversion is not limited to small-scale elements of the graph. The entire graph might be labelled or summarized by some natural language description, as in graph classification \citep{zhang2022topic}. A knowledge graph completion algorithm relying on a transformation of paths in the graph was presented by \cite{lin2023fusing}. In their approach, knowledge graph paths are translated to text by concatenating entities and relations separated by special tokens. Another approach using traversals knowledge graph traversals is OpenDialKG \citep{moon-etal-2019-opendialkg}, which relates shifts in dialogue context to walks in the graph. Other work using this dataset and the dataset of \cite{eric-etal-2021-multi} is the DiffKG model of \cite{tuan-etal-2021-learning}. The DiffKG model uses a transformer architecture to perform walks over the knowledge graph to select KG knowledge for a dialogue system. This system was evaluated with respect to ground truth paths and entities in fixed knowledge bases which are not available in all dialogue domains. The authors explored a baseline following the model of \cite{beygi2022logical}, which transformed a knowledge graph into a text document of semantic  ``pseudo-language" statements. In both this model and DiffKG, a structured text representation of the graph is a source of information to the dialogue system similar to the semantic triple-based model we employ in our study (see Section \ref{sec:experiments}). The AttnIO model of \cite{jung-etal-2020-attnio} also makes use of traversals with an attention mechanism to attend to varying sizes of node neighborhoods in the graph.

\subsection{Large Language Models (LLMs)}
\label{llms_dialogue}

\subsubsection{Response Generation with LLMs}

LLMs have demonstrated strong capabilities for few-shot learning across tasks \citep{brown2020language}, although limitations remain in their use in task-oriented dialogue \citep{hudevcek2023llms}. At the outset, these models have a great deal of information available from the large quantities of data consumed in training them \citep{bommasani2021opportunities}. 

Despite substantial improvements in LLM performance across tasks, limitations remain. \cite{dziri2023faith} investigated LLM performance on reasoning tasks and found that the models often resort to basic pattern recognition rather than generic problem solving. Moreover, \emph{hallucinations} remain a substantial limitation and subject of continuing research \citep{ji2023survey}. As \cite{zhang2023siren} describe it, a hallucination exists when an LLM generates output that ``misaligns with established world knowledge." For our purposes, we can consider established world knowledge to be defined by the physical environment in which the user and robot operate. Thus, the notion of factuality (or, absence of hallucinations) derives from what is true in the physical environment, thus for our study we seek to evaluate factuality of model responses with respect to the environment rather than the graph itself. Recent study has also indicated that the linguistic structures within prompts may be a factor in the occurrence of hallucinations, specifically that more formal and concrete prompts may assist in reducing hallucination \citep{rawte2023exploring}. In this context, evaluation of factuality in LLM responses remains an important direction for study.

\subsubsection{Prompting methods}

In many recent dialogue systems, the model is initialized with a system prompt which consists of an initial input that informs the system what role it should play in the dialogue and how to proceed \citep{liu2023pre}. Besides defining the system's role in the dialogue, the prompt may include other information that is salient to the system's tasks. Importantly, the input to LLMs is bound by a context window beyond which input text must be truncated, although some research has investigated means of alleviating this limitation \citep{ratner2023parallel}. When the total scenario-specific information exceeds the limit imposed by the context window, it becomes advantageous to design methods to select \emph{relevant} information \citep{thulke2021efficient}. Complementary to these methods is the possibility of providing further enriched context such as information derived by commonsense or logical reasoning \citep{walker2023retrieval}.

Due to the difficulty of training large LMs, the use of in-context learning has also become a heavily investigated method of providing training examples to the largest language models \citep{dong2022survey}. With in-context learning, a small number of examples can be used to prime a model to generate output of a desired format or style. Notably, this technique avoids the need to update model parameters which would otherwise be costly or infeasible given the size of many current LLMs, instead relying upon the model's pattern recognition abilities to adapt its output to a desired outcome. Combined with the general notion of prompting, this approach enables substantial flexibility when using LLMs for different tasks and scenarios, including with multimodal tasks such as generalizing models to unseen contexts \citep{tai2023link}. In-context learning with LLMs has also been demonstrated with table-like structures \citep{chen2022large}.

\subsubsection{LLMs in embodied systems}

As noted previously, LLMs have been used as a means of processing language input for embodied systems such as commands or instructions from a user. An example of a model that integrated an LLM with an embodied agent is the ProgPrompt model presented by \cite{gupta2023visual}. ProgPrompt generated task plans for a robot using an LLM by making use of a Python-like prompt format which induced the LLM to generate actions structured as executable programs. Thus, the LLM does not operate on strictly natural language, but rather a kind of programming language. A major advantage in the ProgPrompt system was that the set of actions available to the robot could be represented compactly and abstractly as functions in this language structure. A robot equipped with an LLM was also presented by \cite{billing2023language}, who demonstrated a basic model of HRI with a Pepper robot using the GPT-3 model of OpenAI. In this case, although the dialogue system interacted with the user through an physical agent, the dialogues themselves were not situated in the physical environment. 

Beyond the use of text input (natural language or otherwise), other research has investigated directly incorporating multimodal input to LLM architectures. An example of LLM capable of generating text based on combined text and visual input is the Flamingo model of \cite{alayrac2022flamingo}, who presented an approach to combine textual and image data with few-shot prompting for visual dialogue. Similarly, the Palm-E model of \cite{driess2023palm} addressed multimodality by directly incorporating embeddings of image and other non-language data in the input to an LLM architecture. Such approaches suggest that models trained or prompted to incorporate mixed forms of information can be effective, however these architectures are often task-specific. Moreover, as noted before, it is often cost-prohibitive to train large models.



When the system must take into account its physical environment, the challenge of presenting it appropriate information is potentially greater than a non-embodied dialogue system. As direct operation on audio-visual data or haptic feedback is impossible for an LLM, a transformation from the raw forms of these inputs into natural language is required. \cite{ahn2022can} presented a model called SayCan that combined an LLM with an ``affordance function" describing the probability of success over a set of possible actions for the agent. Another example of such a system was presented by \cite{zhao2023chat}, where a robot ``chats" with the environment in the sense that multimodal sensory input is iteratively presented to an LLM as natural language statements describing the input, with the model outputting actions at each turn. This approach takes advantage of the turn-taking nature of dialogue to inform a system in real-time, but would be difficult to port to an embodied agent intended to explore a wider environment. Nonetheless, this work demonstrates the capabilities of an LLM in reasoning over diverse sources of input information represented as natural language. 

A further example of LLMs integrated in HRI systems is the LLM-Planner of \cite{song2022llm}, a few-shot system using prompting with a GPT-3 model as a planner for an embodied agent. An approach for data augmentation by grounding dialogue responses in text documents was presented by \cite{wu-etal-2020-gcdst}. This model used two dialogue agents to simulate conversations across different domains represented by sets of documents. Lastly, a question answering model which stands quite close to our approach was presented by \cite{lanchantin2023data}, whose model made use of a text-formatted version of a knowledge graph along with a transformer operating directly on the graph. This model operated in a simulated three-dimensional space, where the queries directed to the agent related to the spatio-temporal information describing the user's position in the simulated space. This work did not however investigate the use of a natural language transformation of the graph, relying instead on a text document constructed of structured representations of graph information in a similar manner to the semantic triples model employed in our experiments. Additionally, this approach relied upon the generation of simulated data in a virtual space, whereas we investigate the real-time creation of graph in a physically situated HRI scenario.


\section{Approach}
\label{sec:approach}

In this section, we describe our proposed approach to LLM-based response generation for HRI. Our proposed model revolves around a dynamic knowledge graph that represents both elements of the environment and dialogue utterances. Nodes and edges describing these elements are iteratively added as new data is observed by an agent.

\subsection{State Representation}

As explored in Section \ref{sec:background}, embodied dialogue agents must take into account diverse types of information regarding the dialogue as well as the physical environment. These aspects of embodied dialogue suggest that the flexibility, expressiveness, and semantic interpretability of graphs are effective at representing evolving spatial, temporal, and probabilistic information in a single structure. For these reasons, we use a \textit{dynamic} knowledge graph to represent the dialogue state. 

A dynamic, in-memory knowledge graph permits continuous updates as new information is recorded. The nodes of this graph represent \textit{entities}, in the sense of distinct objects and places in the environments, as well as natural language utterances. Likewise, the edges represent semantic relations between these entities. These relations include spatial or temporal information which may be updated in the course of exploring an environment. Temporality is also applicable to utterances within the dialogue as a chain of responses. As both dialogue information and situated information from the environment are represented within the graph, a form of ``grounding" (in the sense discussed in Section \ref{sec:background}) of the dialogue in the environment is equally possible through edges between utterances and physical objects. 


\begin{figure}[t]
    \centering
    \includegraphics[width=.75\textwidth]{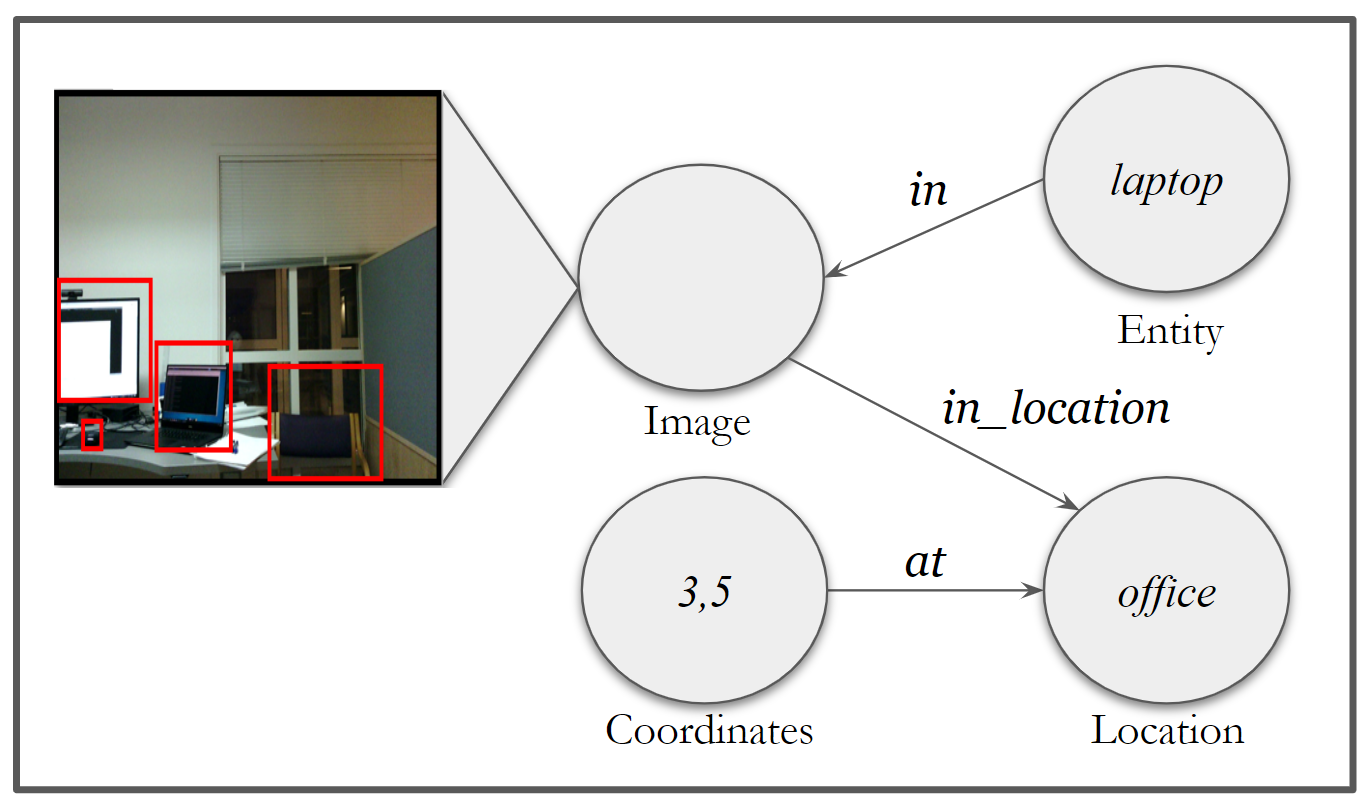}
    \caption{Example of subgraph where an image is assigned a location, and a ``laptop" entity detected in the image data is created as a node with an ``in" relation to the image.}
    \label{fig:graph-image}
\end{figure}

In this model, nodes in the graph are assigned a \emph{type} label along with a \emph{content}. The type of the node indicates what kind of conversational entity the node represents, such as an utterance or a physical object. Edges in the graph represent semantic relations between the nodes of objects, such as physical location, origination, or in the case of utterances a temporal ordering. 

To illustrate situated information in a graph, the node representing a laptop in Figure \ref{fig:graph-image} has a description of ``laptop" along with the type label ``entity". The laptop is likewise related with an ``in" relation to an image data node carrying the type label ``Image", where binary data of the image itself is stored. In turn, the image node relates to a ``Location" node named ``office", indicating that the picture was taken in the location labelled ``office". Plainly stated, this graph structure describes the fact that a laptop computer was seen in an image taken in an office. 

Nodes and edges are also assigned a \textit{probability}. The assignment of a probability value to a node allows for the expression of uncertainty arising from sensory data. This is notably the case for visual elements detected in the images captured by the agent, or for alternative transcriptions produced by the ASR model. In general, regardless of the source or type of information represented in a node, uncertainty in the data can be expressed as an attribute of the node. Likewise, edges can also be assigned to a probability.

A further property of nodes is a \textit{timestamp}. This attribute makes it possible to account for temporality in the data. Every node in our model is assigned a timestamp representing its creation time in the graph. A second timestamp representing the creation time of the \textit{information} is also assigned to the node metadata. This timestamp is the time the information was recorded by an external module such as robot sensors. The difference between the two timestamps is often small, nevertheless separate representations allow precise measurement of any latency or reference to exact occurrences in physical space as opposed to graph update time.

\subsection{State Tracking \& Postprocessing}
\label{sec:statetracking}
\subsubsection{Graph updates}

Dialogue State Tracking \citep{williams2016dialog,ren2018towards} is often viewed in the form of a ``slot-filling" paradigm, wherein slots of a defined ontology are filled by a predictive model. Here, we conceive of state-tracking as the management of the conversational knowledge graph. As the knowledge graph is comparatively small, it is stored in-memory and directly updated through Python operations. Because the knowledge graph is dynamic, new nodes representing spatial positions, images, and entities seen during the embodied agent's movement are added to the state. This is a constant stream of information that updates the dialogue state as soon as new information is obtained. Thus, spatial coordinates and visual information can be continuously added as nodes to the graph while an embodied agent traverses physical space. Equally, language input is added to the graph as it is received, whether concurrently with agent movement or not. This flexibility is an advantage for an embodied agent to allow it to attend to numerous, diverse sources of information simultaneously while also taking into account interactions between them. 

We rely on a dedicated management module that serves as a central point of access for all updates to the graph. Upon each update, the management module triggers all system modules that should be notified of this change. Those modules may in turn lead to new graph updates, thereby creating a cascading transformation to the graph, where modules variously react and push changes to the graph manager. An important example is a response generation module, which is triggered upon the addition of a new user utterance node in the graph. Using the updated graph, the response generation module generates an agent response to the last user utterance and adds the response to the graph with a ``follows" relation to the utterance. This graph update in turn is passed to the embodied agent for speech generation.

In the same fashion, objects detected in visual sensory input are added to the graph. Raw image data collected by a camera can be added as nodes to the graph directly. These additions in turn trigger postprocessing of the image data, whereby objects, people and other entities may be recognized and added to the graph as additional nodes with an ``in" relation to the image. We also assign probabilities to these objects derived from the output of the visual model use to identify them.


\begin{figure}[t]
    \centering
    \includegraphics[width=.80\textwidth]{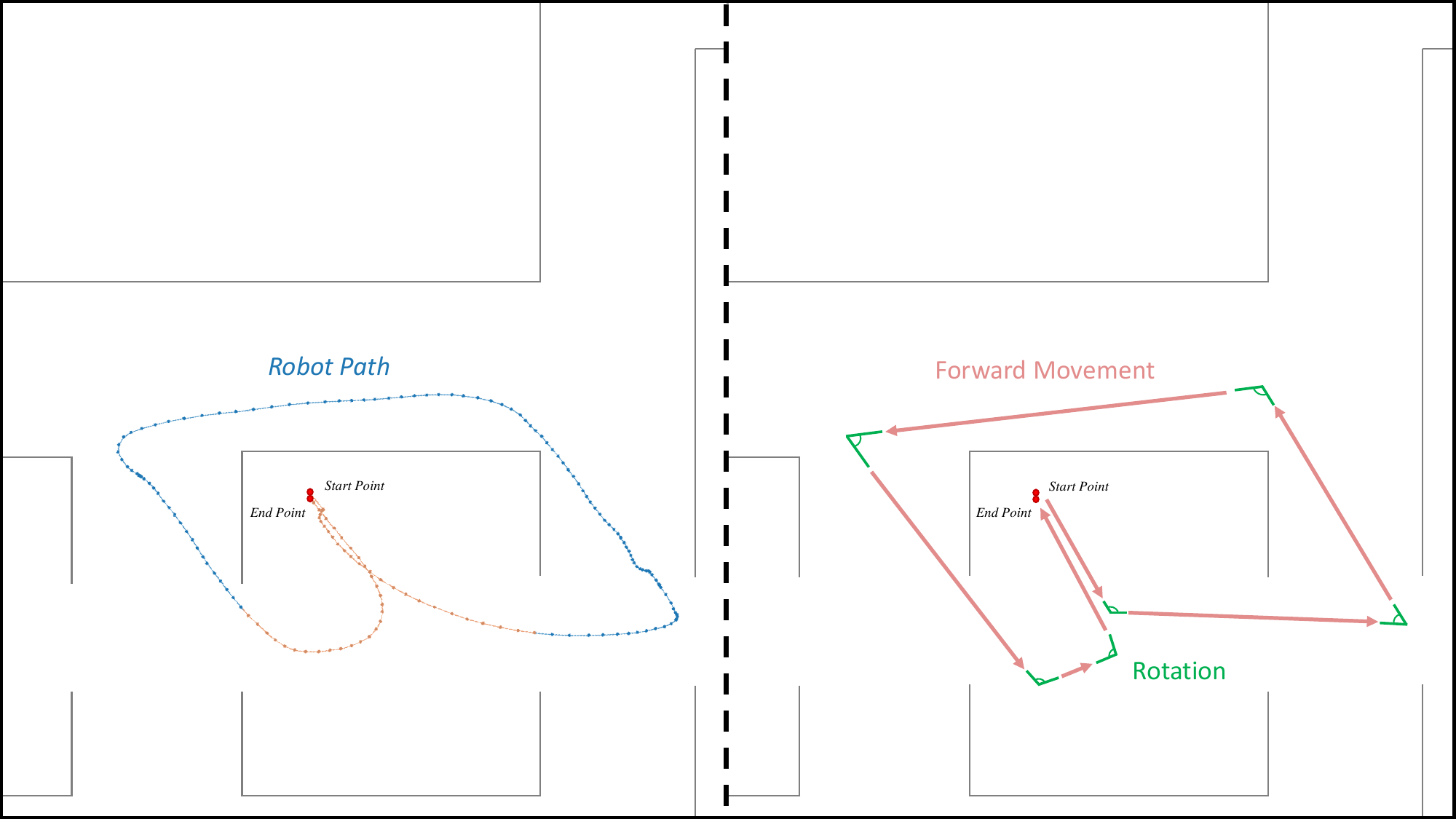}
    \caption{Example path drawn from spatial coordinates, with raw coordinates on the left and the approximated path on the right. The orange points mark coordinates associated with the room label, blue points are those labeled as located in the hallway. Approximate forward movements and in-place rotations are highlighted alongside the robot path in red and green, respectively.}
    \label{fig:pathmap}
\end{figure}

\subsubsection{Path generation}

A particular component integrated in the HRI platform employed in our experiments (Section \ref{sec:experiments}) is a mechanism for deriving a continuous path from the raw sequence of spatial coordinates recorded by the robot. This mechanism operates in two steps. The spatial coordinates are first converted into a limited number of lines based on the Ramer-Douglas-Peucker algorithm \citep{douglas1973algorithms}. Those lines are in turn converted into an ordered list of movements, divided in two types: \emph{straight forward movements} and \emph{in-place rotations}\footnote{This is of course a simplification, based on the fact that the Pepper robot uses omni-directional wheels, and that its movements in the indoor environment of the user study were either straight-line motions or in-place turns.}, the straight movements being specified for a given distance, and the rotations for a given angle. This list of movements is then integrated into the graph. An example path drawn from raw coordinates is given in Figure \ref{fig:pathmap}.

The nodes expressing the spatial coordinates in the state graph are further connected to edges expressing that they are part of a larger \emph{location} with a given name, such as ``hallway'' or ``break room''. In the experiments, those names are provided by the human participant during the tour, as detailed in Section \ref{subsec:setup}.


\subsection{Graph-to-text Transformation}
\label{sec:verb}

\subsubsection{Verbalization function}

\begin{figure}[t]
    \centering
    \includegraphics[width=1\textwidth]{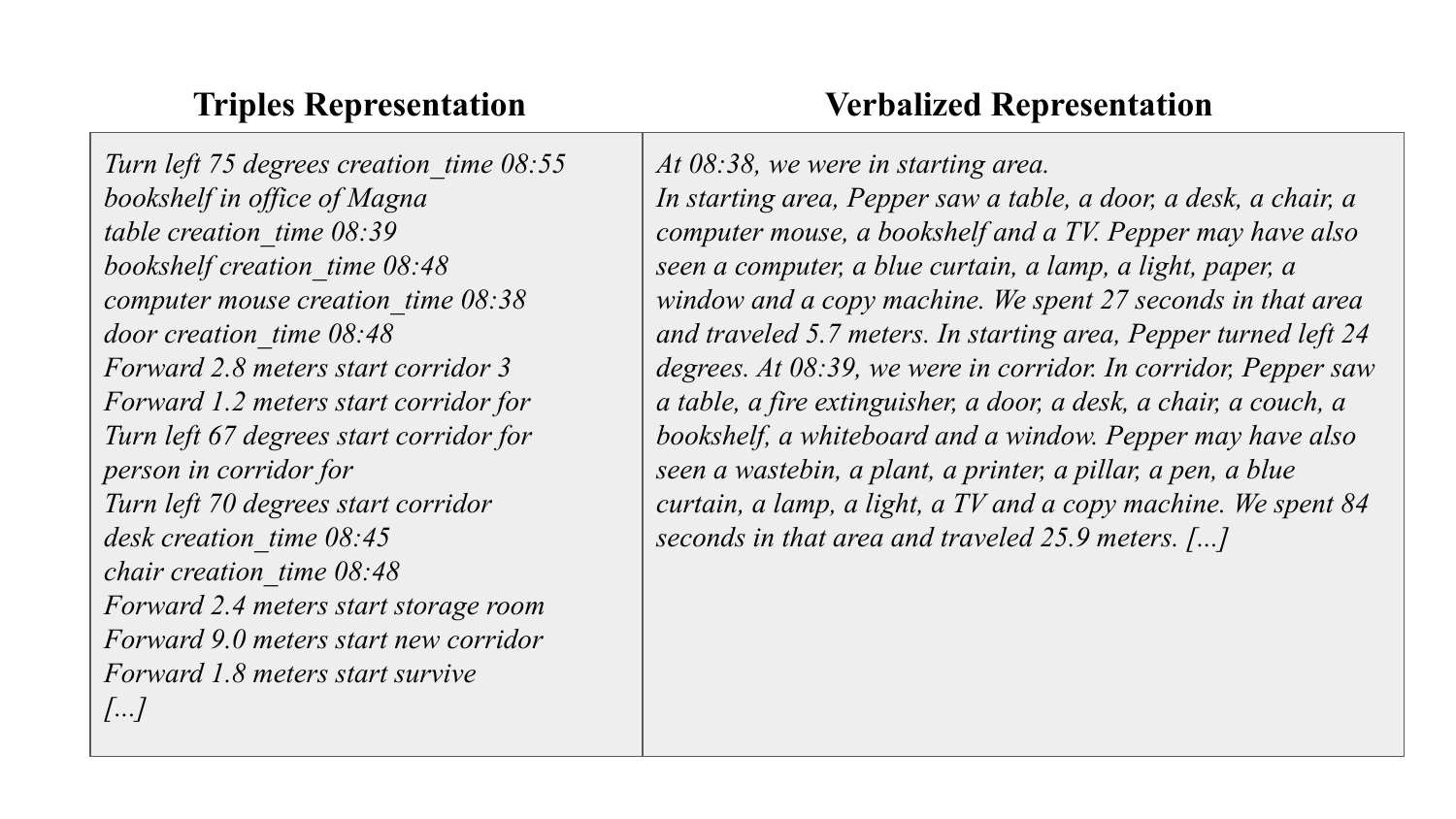} \vspace{-8mm}
    \caption{Comparison of a dialogue graph represented with triples versus verbalization.}
    \label{fig:verbalization}
\end{figure}

To use the dialogue state graph information with an LLM, it is necessary to convert the structured information into natural language. A variety of methods have been applied to generate natural language from structured information such as tables \citep{nema-etal-2018-generating}. In our proposed approach, the information expressed in the dialogue state graph is converted into text description via a set of \emph{graph traversals}. In this approach, both relations between nodes as well as more complex paths can be rendered with a natural language description. This traversal-based transformation is a simple means to convert structured information to a form amenable to an LLM. 

To traverse the graph, we first order the nodes temporally and iterate over them. For each node, the node's type determines which paths should be searched for in its neighborhood. The traversal function then collects ingoing and outgoing edges along with the neighboring nodes connected to them. For each node type, the function defines a set of paths to traverse by searching along the ingoing or outgoing edges. As an example, when evaluating an ``entity" node, paths containing an ``in" relation from the source node to an image node with a subsequent ``in\_location" edge to a ``location" node describes the location of an object (as in Figure \ref{fig:graph-image}). Thus, the paths are defined in terms of which edges and node types should be found in a given sequence of hops in the graph. The collection of these paths then constitutes a chronologically ordered description of the graph. Node probabilities are also recorded during the traversal.

The verbalization process takes the form of a function that converts the graph into a string according to a set of handcrafted \emph{parameters}. An example parameter is that the system may reference itself within the prompt, such as first or third person pronouns, or a name. Depending on this parameter, the description of the graph will then summarize the information from either a personal or impersonal perspective with respect to the dialogue agent. As illustrated in Table \ref{tab:parameters}, the verbalization function employed in our experiments also included parameters related to the use of discourse markers, the inclusion of low-probability entities in the description, or how to express distances, angles, and time. 


\subsubsection{Parameter selection}

As mentioned above, the transformation of the graph to a text document is a parameterized function. To select the best set of parameters for the task, we optimize the parameters with respect to the output of a LLaMA language model \citep{touvron2023llama}, such that the model assigns the highest probability to the correct response out of a set of candidate responses across dialogues. For this purpose, we collect a small set of Wizard-of-Oz interactions, each response from the wizard being associated with a corresponding dialogue state graphs (which include sample locations and entities along with the dialogue history). 

Formally, let $V_{\mathbf{P}}(G)$ denote a graph-to-text function where $G$ is a knowledge graph and $\mathbf{P}$ a set of categorical parameters, and let $D = \{(G_i, r_i), 1 \leq i \leq N\}$ denote the recorded set of $N$ Wizard-of-Oz responses $r_i$ in a conversational context represented by its dialogue state graph $G_i$. We then search for an optimal set of parameter values $\mathbf{P}^*$ that minimizes the cross-entropy loss over the Wizard-of-Oz examples, assuming a fixed language model $\textrm{LLM}$:
\begin{equation}
    \mathbf{P}^* = \argmin_\mathbf{P} \mathcal{L}_{\textrm{CE}} (D \ ; \textrm{LLM}, \mathbf{P})
\end{equation}

where the cross-entropy loss is itself defined as:
\begin{equation}
    \mathcal{L}_{\textrm{CE}} (D \ ; \textrm{LLM}, \mathbf{P}) = - \frac{1}{N} \sum_{(G_i, r_i) \in D} \log P_{\textrm{LLM}} (r_i \ | \  V_{\mathbf{P}}(G_i))
\end{equation}

The optimization is performed using a tree-structured Parzen estimator algorithm \citep{bergstra2011algorithms}, as implemented in the Optuna package for efficient hyper-parameter optimization \citep{optuna_2019}. Using the above approach, we are able to optimize the categorical parameters and adjust the verbalization function with only a few example responses from a wizard. 

\begin{table}[t]
    \centering
    \begin{tabular}{lr}
    \toprule
       Parameter  &  Values \\
       \midrule
       How the agent refers to itself  & \textbf{``Pepper"}, ``the robot", ``I" \\
       Whether to include discourse markers between sentences & True | \textbf{False} \\
       How to specify distance (in meters) & \textbf{``precise"}, ``rounding", ``none" \\
       Whether to specify rotation angle (in degrees) & \textbf{True} | False \\
       Whether to include low-probability entities in description & \textbf{True} | False \\
       Whether to include time in description & \textbf{True} | False \\
       Whether to mention the total number of turns & True | \textbf{False} \\
       \bottomrule 
    \end{tabular}
    \caption{Parameter values for the verbalization function. Each parameter specifies an aspect of the verbalization. Bold values indicate the parameter values selected by optimization.}
    \label{tab:parameters}
\end{table}

\subsection{Response Generation}

The final component of the proposed approach is the actual response generation. As already mentioned, this response is generated using a large language model optimized for dialogue use cases. In our experiments, those language models were respectively based on GPT-4 \citep{openai2023gpt4} and the Llama-2 chat model \citep{touvron2023llama}. 

The response is generated based on a \emph{prompt} that includes the verbalized graph along with:
\begin{itemize}
    \item A short, generic instruction text that details how the robot should respond to the user utterances, and stress in particular that the robot responses should all be grounded in the observations expressed in the verbalized graph. 
    \item A short list of example responses from the Wizard-of-Oz interactions to take advantage of in-context learning \citep{brown2020language,wei2022emergent} and provide cues about how responses are expected to be formulated.
\end{itemize}

Once decoding is complete, the response is itself added to the graph as a new node. This node is connected to the user utterance with a \texttt{responds\_to} relation, thereby capturing the temporal sequence of dialogue turns in the graph.

\section{Experiments}
\label{sec:experiments}

In this section, we describe  the design of our experimental study and the results of the corresponding user evaluation. The goal of our study is to explore the capabilities of a robot that relies on a dynamic knowledge graph as the representation of its environment and dialogue state, and converts this graph into natural language to enable the generation of system responses through an LLM. To do so, we designed a study wherein users brought the robot to different locations in an office environment. After the completion of this office tour, the participants were then instructed to conduct a short dialogue with the robot to assess its ability to recollect the observations made during the tour.

Three distinct model setups were investigated, and users conversed with all three in a randomized order immediately after their tour with the robot. After each conversation, the participants were asked to rate the robot responses with respect to their factuality and their adequacy.  

We recruited a total of 20 participants (for the most part IT students and researchers) for our study, the first two participants being asked to test a prototype version of the final system. Each participant was given a 200 NOK gift card for their participation. For the robot in the experiment, we use the Pepper robot of Aldebaran\footnote{\url{https://www.aldebaran.com/en/pepper}}. The dialogues were conducted in English.


\begin{figure}[t]
    \centering
    \includegraphics[width=.75\textwidth]{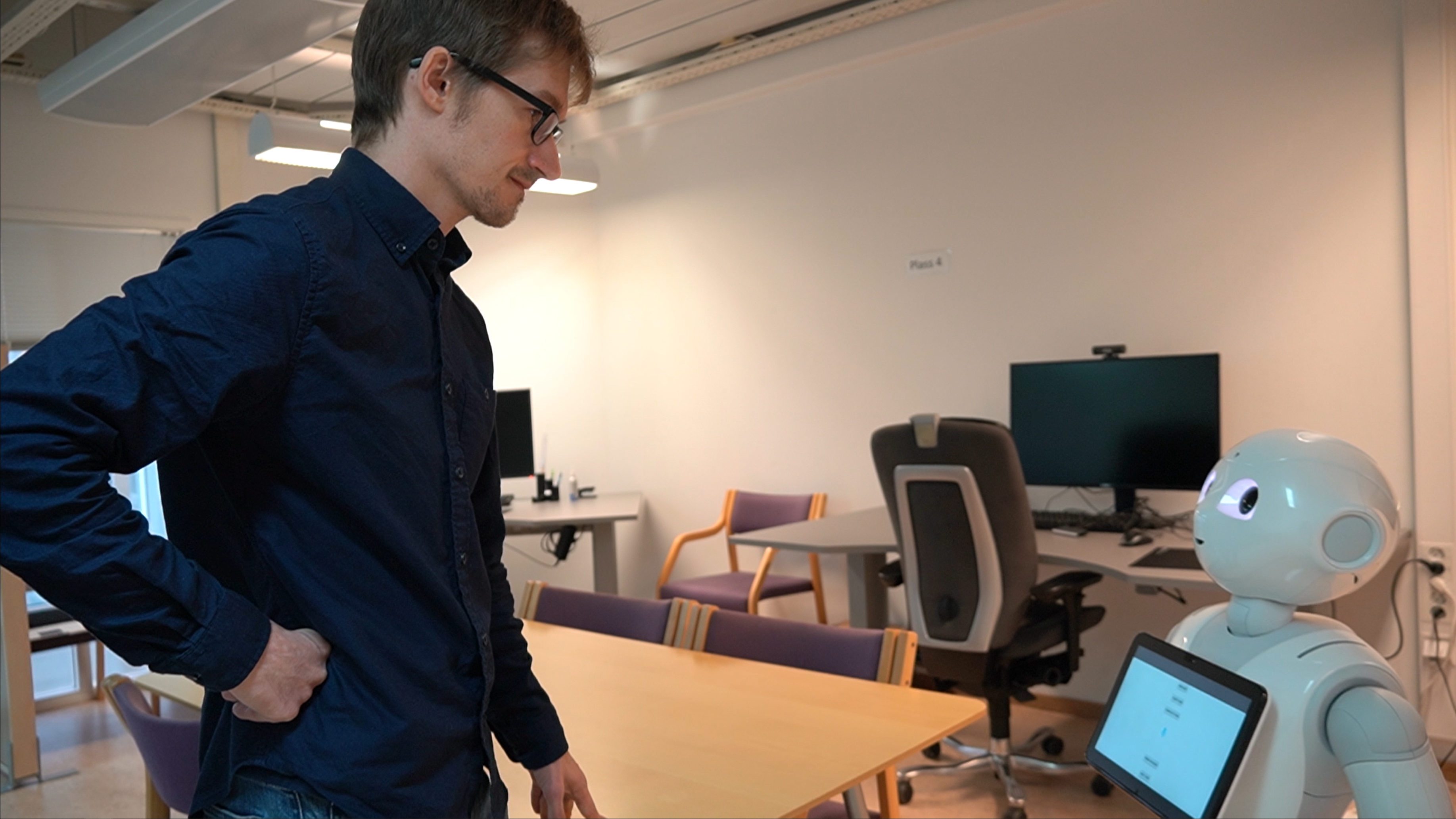}
    \caption{The first author in a dialogue with the robot. A video of a short example tour may be found at \url{https://youtu.be/a52zBcfVgS8}}
    \label{fig:robot}
\end{figure}

\subsection{Experimental Setup}
\label{subsec:setup}

The experiment was structured in two parts that respectively correspond to the physical tour of the office and the dialogue between the user and the robot that followed. 

\paragraph{Tour phase} During the tour, the robot is moved around an office environment by a participant to record details of its path, locations, and identifiable objects. For the sake of simplicity, we instruct participants to push the robot along its path in the office. Before entering a new room or corridor, the participant was instructed to activate the robot microphone by tapping its head and stating the name of the area they were entering. 

The time spent on each tour ranged from around 5 to 20 minutes.

\paragraph{Dialogue phase} After the tour was completed, the user was then asked to converse with the robot and have it summarize the tour, as illustrated in Figure \ref{fig:robot}. The participant was instructed to ask the robot at least 5 questions relating to the tour, asking it to summarize the tour along with questions such as to describe which locations it visited, what was seen in each location, what distance was traveled, or how much time was spent in an area. Participants were encouraged to treat the conversation as a natural, continuous dialogue rather than a set of isolated questions, allowing for follow up questions referencing the previous question (e.g. ``What did you see after that?"). 

This dialogue between robot and user is repeated three times, each time with a distinct configuration for response generation (see below). Participants were instructed to treat each dialogue as a completely independent conversation. The model order was randomly reshuffled for each experiment. The participants were also told they did not need to ask the same questions in each dialogue.





\subsection{System configuration}

\begin{figure}[t]
    \centering
    \includegraphics[width=.75\textwidth]{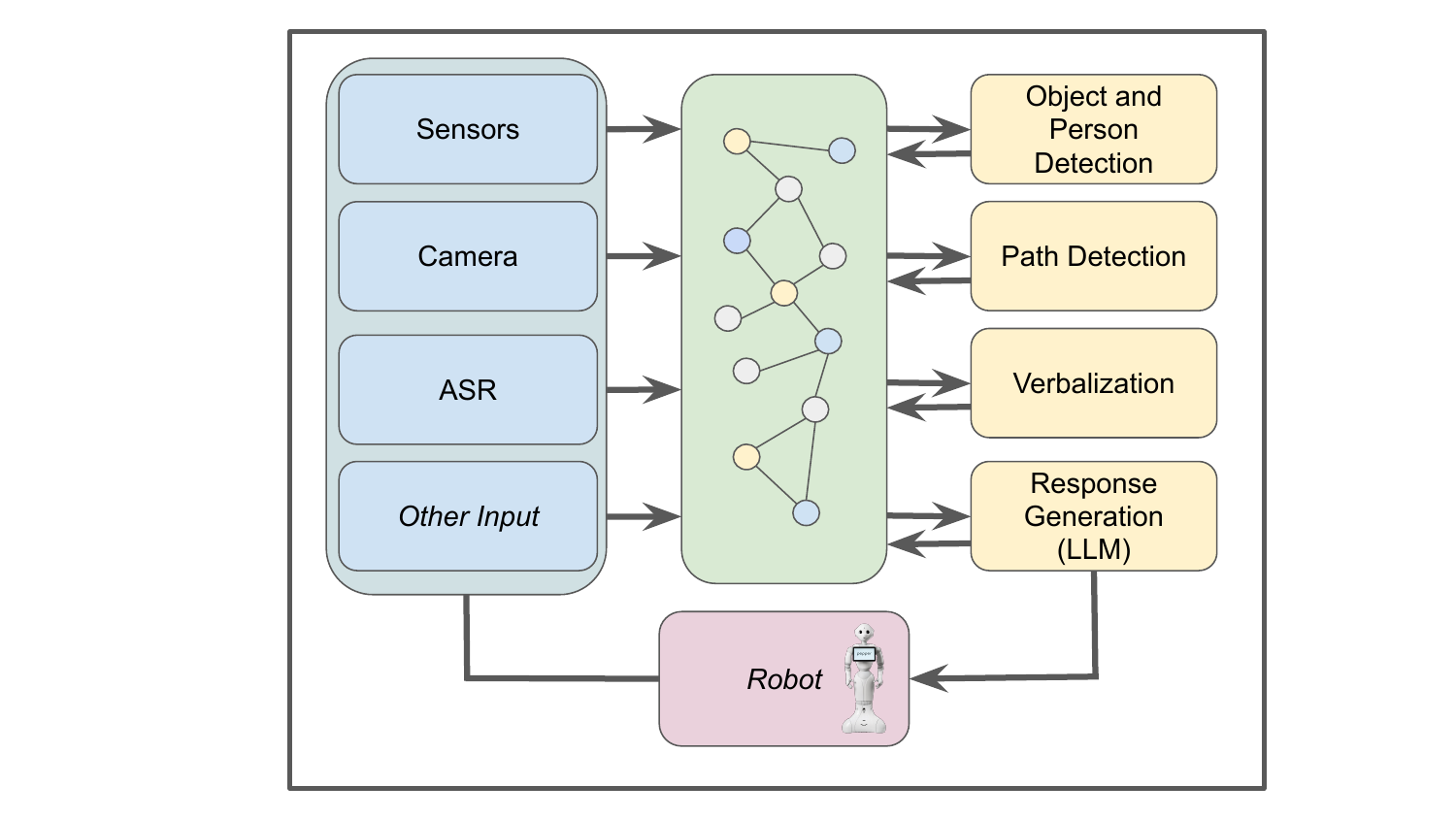}
    \caption{Diagram of the implemented architecture for the experiment. Robot sensor modules are colored blue, while post-processing models operating on the graph are colored yellow.}
    \label{fig:architecture}
\end{figure}

The software architecture employed for the experiments integrates a number of modules, as illustrated in Figure \ref{fig:architecture}. The sensor-related modules ran on the Pepper robot itself, while the other modules were located on a separate machine with access to a GPU. 

\subsubsection{Robot perception}

During the tour, the dialogue state graph is continuously updated with new visual and positional data, along with the timestamp at which these elements are observed by the robot. The user-provided location labels are added to the graph as nodes with the provided name as the node description. As the robot moves along its path, new nodes are added to the state representing spatial positions, images, and entities seen during its movement. The room labels provided by the user are assigned to all subsequent positions recorded by the robot until the next label was provided.

The robot takes a picture of what is in front of it approximately every 1.5 seconds, along with its current coordinates relative to an initial starting location. Due to odometry drift, we were unable to treat the spatial data as a precise representation of the robot's path at large distances. Nevertheless, these location measurements provided an approximate measure of distance traveled within areas as well as a means of estimating directional changes by the robot. The image data and $x,y$ coordinates are added to the graph as nodes labeled  \texttt{image} and \texttt{position}, respectively, and are each created with an edge to the current location node. As mentioned in Section \ref{sec:statetracking}, a dedicated mechanism is implemented to convert the raw spatial coordinates to a high-level sequence of movements based on the Ramer-Douglas-Peucker algorithm \citep{douglas1973algorithms}.

When an image is added to the graph, a visual question answering (VQA) model processes the image information and returns a score of whether objects were detected in the image. The model we chose for this purpose is the  Vision Transformer model of \cite{minderer2022simple}. We enumerate the possible objects the robot should see including people as well as office objects (see Appendix \ref{sec:appendixA}). When objects are added to the graph with this model, the estimated probability of the object in the image from the VQA model is assigned as the probability of the node. As this is based upon a non-fine-tuned model, there is substantial noise in the observations of the objects.

\subsubsection{Dialogue system}

During the dialogue phase, the user activates the microphone by tapping the robot's head. Utterances are transcribed using the Google Speech API\footnote{\url{https://cloud.google.com/speech-to-text}} and insert to the knowledge graph as nodes. 

The process of response generation is triggered upon the addition of new utterance nodes in the graph. This process begins with the verbalization of the graph (see section \ref{sec:verb}). The resulting description of the graph is then inserted into a prompt format for the language model.  We use GPT-4 \citep{openai2023gpt4} and LLaMA 2 \citep{touvron2023llama} as the LLMs for our experiment. The GPT-4 model was accessed through the OpenAI API while the Llama-2 model, which is specifically optimized for dialogue use cases\footnote{\url{https://huggingface.co/meta-llama/Llama-2-13b-chat-hf}}, was run on a local GPU, using quantization methods \citep{dettmers2022optimizers,dettmers2022llmint8} to satisfy memory limitations. 

To evaluate the impact of the graph verbalization on the dialogues, we test the following setups with these models:
\begin{enumerate}
    \item The GPT-4 model provided with a prompt containing the verbalized dialogue state graph.
    \item The LLaMA 2 model provided with a prompt containing the verbalized dialogue state graph.
    \item The GPT-4 model provided with a prompt containing the dialogue state graph expressed as semantic triples of the state graph facts.
\end{enumerate}


The first two models make use of the verbalized graph as produced by the transformation function and selected parameters from Section \ref{sec:verb}. The prompt to these models is therefore a concatenation of the general prompt describing the system's role and the resulting text description. For the third model, the graph information is instead provided as a list of semantic triples directly describing the edges of the graph, as illustrated in Figure \ref{fig:verbalization}.  

Due to technical issues encountered with a few interactions, there is a slight imbalance in the number of dialogues collected for each model (GPT-triples: 16, GPT-Verbal: 17, LLaMA: 19). This, however, does not affect the statistical significance of the experimental results, detailed below. 

The LLM-generated robot responses are sent to the text-to-speech engine embedded on the Pepper platform. When speaking, the robot also makes basic hand gestures and movement to accompany its response, using the off-the-shelf library made available on the Pepper platform. 

\subsection{User Evaluation}


Right after the conclusion of each dialogue, the participants were asked to rate two statements about the dialogue on a Likert scale of 1 to 5 for two statements where 1 is ``Never", 2 is ``Mostly Not", 3 ``Sometimes", 4 ``Mostly", and 5 ``Always":
\begin{description}
    \item[\textbf{Statement 1 (factuality)}:] The robot responded \textit{factually} according to what we saw on the tour.
    \item[\textbf{Statement 2 (conversational adequacy)}:] The robot answered my questions and responded in a \textit{relevant}, \textit{natural}, and \textit{concise} manner.
\end{description}

\begin{table}[t]
    \begin{center}
        \begin{tabular}{lcc} \toprule
            \textbf{Model} & \textbf{Factuality} & \textbf{Adequacy} \\ \midrule
            {$\textrm{GPT}_{\textrm{Triples}}$} & 3.44 & 3.69 \\
            {$\textrm{GPT}_{\textrm{Verbal}}$} & 4.06 & 3.88 \\
            {$\textrm{LLaMA}_{\textrm{Verbal}}$} & 3.68 & 3.79 \\ 
            \bottomrule 
        \end{tabular}
        \caption{Average participant scores for the model factuality and conversational adequacy.}
        \label{table:participant_ratings}
    \end{center}
\end{table}

\begin{figure}[h]
    \centering
    \includegraphics[width=1\textwidth]{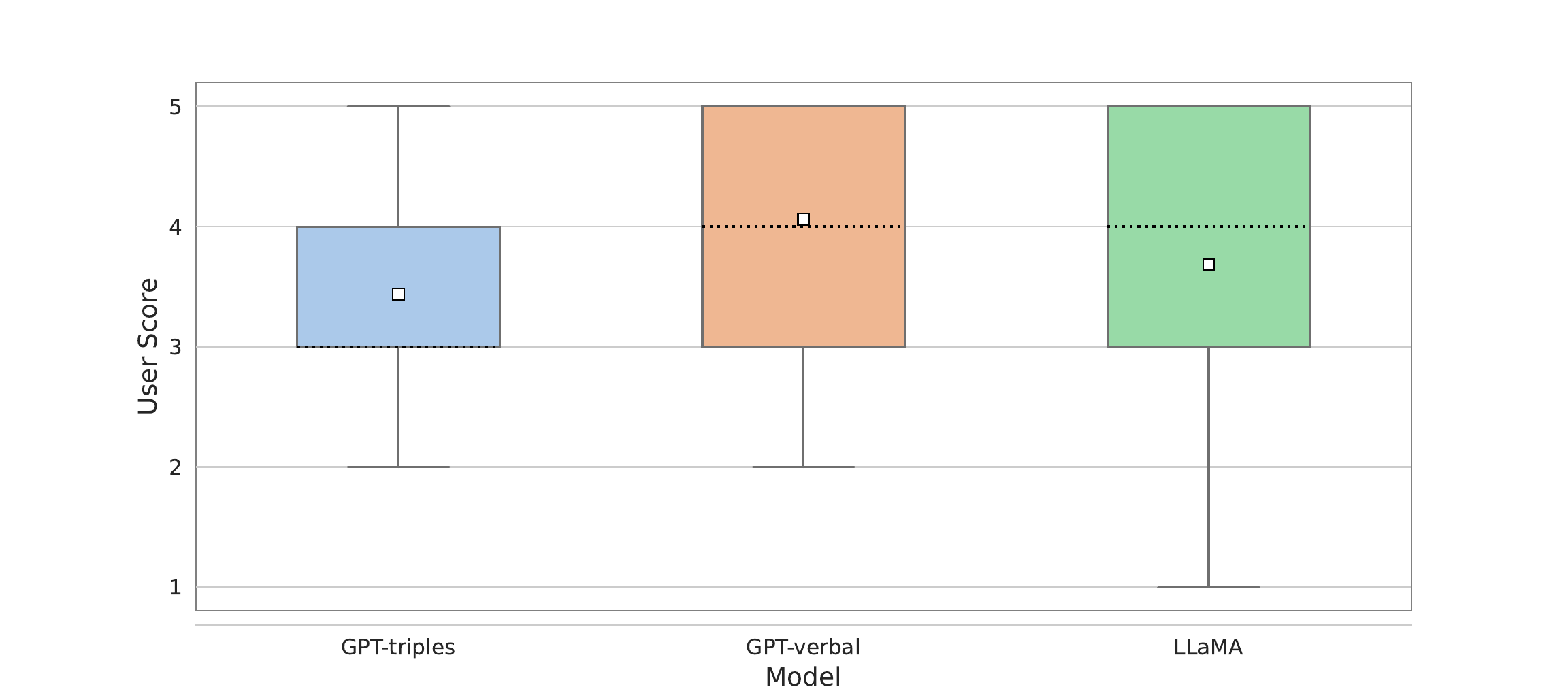}
    \caption{Box plot of the results for Statement 1: The dotted line in each box represents the median, while the white inner square displays the mean user score for each model.}
    \label{fig:boxplot}
\end{figure}

Table \ref{table:participant_ratings} shows the average scores given to each model for both statements, while Figure \ref{fig:boxplot} provides a box plot of the user scores for Statement 1. 
The user ratings show that participants generally rated the models using verbalized documents higher in factuality than the model based on semantic triples, while providing similar scores to all models for the second statement regarding the conversational adequacy of the robot responses. 

To statistically evaluate the difference in scores between the models, we use the Wilcoxon Signed Rank Test \citep{wilcoxon1945individual} with the typical threshold for significance $\alpha=0.05$. We first evaluate both alternative models pairwise against the baseline model. Regarding the first statement, between the GPT-verbal model and the GPT model based on semantic triples we find a difference of $p=0.03, t=1.86$. We found no statistically significant difference between the scores of the LlaMA model and the semantic triples model ($p=0.13$, $t=1.13$) for this statement. Likewise, we find no statistically significant difference between either GPT-verbal ($p=0.25$, $t=0.67$) or LLaMA ($p=0.29, t=0.54$) scores with the semantic triples model for the second statement. 

While the pairwise difference in distributions of scores for factuality between GPT-Verbal and the baseline were below our threshold $\alpha=0.05$, the experiment results may also be looked at from the perspective of \textit{all} models. That is, whether for all models we can say there is a statistically significant difference in outcomes (whatever model is chosen of the three). To correct for multiple comparisons of this sort, we apply Bonferroni correction to provide a conservative bound on the resulting p-values amongst all model comparisons. With Bonferroni correction, between the GPT-verbal model and the baseline triples model the adjusted value for factuality is $p=0.1$, and for adequacy $p=0.75$. For the LLaMA model, the values are $p=0.39$ and $p=0.88$, respectively. 


\subsection{Analysis}

\subsubsection{General observations}

Qualitatively, the majority of factuality errors across the three models were due to objects that were not detected by the VQA model employed for object and person detection, and were consequently not mentioned in the provided prompt (either as semantic triples or verbalized graph). Although those omissions do indeed result in factually incorrect statements from the robot, there is little that can be done from the dialogue side to correct this shortcoming. 

The setup based on semantic triples demonstrates a decent performance on both statements rated by the participants. 
This model was however more prone to asserting it did not have information available, and was generally unable to provide details as to the order of rooms visited. Although the times rooms were entered were available to the model based on semantic triples, the structure of the input made it difficult for the LLM to provide informed answers over the tour chronology. 


Due to limited GPU resources, the LLaMA model had much higher latency in response time to user utterances. This condition did not appear to affect the resultant scores, but with inspection of the dialogues it appears to have resulted in a higher number of independent questions rather than ones that build upon each other. Participants generally appeared to feel more free to fluently express themselves with the GPT models, sometimes to the point of off-topic utterances. Moreover, the LLaMA model's responses to questions about what it saw tended to lean to absolute ``Yes" and ``No", rather than expressing uncertainty. Another tendency of the LLaMA model was to provide numbered lists of observations within the tour, resulting in more verbose conversational behaviour. 

\subsubsection{Negation words}

To further explore the difference in factuality scores, we also analyzed the occurrence of negation words. The following list was used: ``no'', ``not'', ``can't'', ``don't'', ``unable'' and ``cannot''. 

As shown in Table \ref{table:negations}, the number of negation words was much higher in the model based on semantic triples. The higher occurrence of those words reflects the tendency of this model to refrain from providing certain details, particularly the order of visited locations and the time spent in them. 




\begin{table}[h]
    \begin{center}
        \begin{tabular}{lccccc} \toprule
            \textbf{Model} & ``Not" & ``Cannot" & ``No" &  \textit{Other*} & \textbf{Total} \\ \midrule
            {$\textrm{GPT}_{\textrm{Triples}}$} & 69 & 13 & 14 & 16 & 112\\
            {$\textrm{GPT}_{\textrm{Verbal}}$} & 48 & 3 & 2 & 15 & 68\\
            {$\textrm{LLaMA}_{\textrm{Verbal}}$} & 56 & 0 & 28 & 7 & 91\\ 
            \bottomrule 
        \end{tabular}
        \caption{Count of negation words by model.}
        \label{table:negations}
    \end{center}
\end{table}

\subsubsection{Expression of uncertainty}

To investigate the models' ability to express probability and uncertainty, we analyzed a count of words expressing uncertainty such as ``may" and ``might". The frequency of such words reveals that the GPT-4 model with the verbalized graph was substantially more likely to communicate uncertainty to the user, as shown in Table \ref{table:uncertainty_words}. 


\begin{table}[h]
    \begin{center}
        \begin{tabular}{lccccc} \toprule
            \textbf{Model} & ``May" & ``Might" & ``Possibly" &  ``Possible"  & \textbf{Total} \\ \midrule
            {$\textrm{GPT}_{\textrm{Triples}}$} & 4 & 0 & 2 & 0 & 6 \\
            {$\textrm{GPT}_{\textrm{Verbal}}$} & 25 & 7 & 7 & 1 & 40 \\
            {$\textrm{LLaMA}_{\textrm{Verbal}}$} & 7 & 0 & 0 & 1 & 8 \\ 
            \bottomrule 
        \end{tabular}
        \caption{Token counts of words expressing uncertainty by model.}
        \label{table:uncertainty_words}
    \end{center}
\end{table}

Upon inspecting the dialogues, there appear to have been no instances where the GPT-verbal model expressed uncertainty with the word ``may" that was not justified by the observation descriptions. By constrast, The GPT-triples model  used ``may" in this way twice, with the remaining count arising from utterances where the model asks the user to note that the observations may not describe certain aspects such as order or count of objects. This tendency appears to demonstrate that while probability of observations was presented in semantic triple form, the model made significantly more use of uncertainty in its expression when uncertainty was transformed to natural language.

\subsubsection{Influence of model order}

We analyzed the effect of the model order on user scores, shown in Table \ref{table:order_ratings}. There was not a substantial effect on scores from whether the model was presented first or last to the user. To analyze the interaction of model types with order, we fit a Bayesian linear mixed model using the Bambi\footnote{https://bambinos.github.io/bambi/} library for both factuality and adequacy scores. We treat model order as a random effect as $Score \sim Model * (1|Order)$ to evaluate the interaction of model and order on participant scores for each question. The results indicated that order had minimal impact on scores, with a random intercept variance estimate of  0.109 (SD = 0.330, 94\% HDI: [0.000, 0.976]) for factuality and 0.203 (SD = 0.451, 94\% HDI: [0.002, 1.573]) for adequacy.

\begin{table}[h]
    \begin{center}
        \begin{tabular}{lcccc} \toprule
            \textbf{Model} & \textbf{Factuality (first)} & \textbf{Factuality (last)} & \textbf{Adequacy (first)} & \textbf{Adequacy (last)} \\ \midrule
            {$\textrm{GPT}_{\textrm{Triples}}$} & 3.5 & 3.5 & 3.83 & 3.67 \\
            {$\textrm{GPT}_{\textrm{Verbal}}$} & 4.4 & 4 & 3.8 & 3.75 \\
            {$\textrm{LLaMA}_{\textrm{Verbal}}$} & 3.71 & 3.86 & 3.86 & 3.71 \\ 
            \bottomrule 
        \end{tabular}
        \caption{Average participant scores for model factuality and conversational adequacy when presented to the user first and last.}
        \label{table:order_ratings}
    \end{center}
\end{table}

\subsubsection{Outliers}

As shown in the box plot of Figure \ref{fig:boxplot}, the LLaMA model had the widest range in scores, being the only model to receive scores of ``1" to either statement. This score was recorded in two dialogues. In one case, the model frequently answered that it had not seen things in the tour after having enumerated them in an initial summary and demonstrated an unexpected inability to recall facts. This first dialogue is as follows and received a score of 2 for statement 2:
\begin{quote}
    \textbf{User:} which room did we spend the most time in \\
    \textbf{Robot:} Based on the observations, we spent the most time in the "sick room" (205 seconds). \\
    \textbf{User:} what was the room we went to after the glass hallway \\
    \textbf{Robot:} Based on the observations, after the glass hallway, we went to the "printer area" (32 seconds). \\
    \textbf{User:} do you see any useful equipment for extinguishing a fire anywhere \\
    \textbf{Robot:} No, I do not see any useful equipment for extinguishing a fire anywhere in the observations. \\
    \textbf{User:} were there any chairs in the glass hallway \\
    \textbf{Robot:} No, I did not see any chairs in the glass hallway. \\
    \textbf{User:} were there any people working in the office \\
    \textbf{Robot:} Yes, I saw a person working in the office. \\
    \textbf{User:} what furniture store in the sick room \\
    \textbf{Robot:} There was no furniture store in the sick room.
\end{quote}

While the first two observations were correct according to the recorded time, the model was unable to provide facts about the detected entities, namely a fire extinguisher and a chair in the hallway, furniture in the sick room (chair, sofa) and the absence of people in the office.

In the second case, the model failed to describe detected objects from the tour and was given a score of 1 for both statements. However, all model scores given by the participant were quite low. 
This participant appeared to take a much stricter approach to interpretation of the statements in evaluating all models. The model was nonetheless weak in recalling detected objects in this dialogue. An excerpt from this dialogue is as follows:

\begin{quote}
    \textbf{User:} where did we go first \\
    \textbf{Robot:} We first went to the main hallway. \\
    \textbf{User:} and after that \\
    \textbf{Robot:} After the main hallway, we went to the print room. \\
    \textbf{User:} did you see any plants \\
    \textbf{Robot:} No, I did not see any plants during my tour of the office floor. \\
    \textbf{User:} that's not true \\
    \textbf{Robot:} I apologize, you are correct. I did see a bookshelf in the first office I visited, but I did not see any plants.
\end{quote}

Overall, the LLaMA model also demonstrated a propensity to answer particularly short introductory greetings with a list of answers to the example questions provided in the prompt, a factor which may have influenced the scores.



\subsection{Dialogue examples}

Dialogues \ref{table:example_dialogue1} and \ref{table:example_dialogue2} provide two additional examples of interactions, respectively based on the GPT-4 model relying on semantic triples and the GPT-4 model using the verbalized graph. 

As illustrated in Dialogue \ref{table:example_dialogue1}, one peculiarity of the responses based on semantic triples was the difficulty of providing information about seen objects in an area even when recalling it in other locations. 
This kind of error may relate to the model needing to attend to a number of individual facts about the plants in the data at once, thus overlooking facts where the information is available. While the information necessary to determine the chronological ordering of locations visited, time spent in locations, and distances traveled are present in the triple-structured information, the results suggest that it is difficult for the model to fully make use of this information. 

\renewcommand{\figurename}{Dialogue}
\begin{figure}[p]
    \begin{tabular}{p{0.99\linewidth}} \toprule
      \textbf{GPT-triples} \\ \hline
        \textbf{User:} can you summarize our tour \\
        \textbf{Robot:} During our exploration of the office floor, we visited various locations such as corridors, a printer room, storage room, offices, a coffee break area, and a reception area. We found many items such as doors, blue folders, a copy machine, bookshelves, a TV, a light switch, and a wastebin. There were also different movements like turning right or left and proceeding forwards for various meters. I also identified the presence of people, but I cannot confirm their identities. The exploration began and ended at different times as indicated by the creation time of the objects and paths. \\
        \textbf{User:} how many people did we see \\
        \textbf{Robot:} It's unclear exactly how many people we saw because they were often identified simply as `person', `man', or `woman'. These labels could refer to the same or different individuals. Therefore, I cannot accurately provide the total number of people we saw without further information. \\
        \textbf{User:} did we go up to the starting area \\
        \textbf{Robot:} Yes, there were several mentions of moving to a location labeled as the ``starting area" during our exploration. \\
        \textbf{User:} how long did we spend in the coffee room \\
        \textbf{Robot:} The observations do not provide specific information about the amount of time spent in the coffee break area. \\
        \textbf{User:} which area did we visit after the coffee room \\
        \textbf{Robot:} The observations do not provide a sequential order of the locations visited, so it's unclear which area was visited immediately after the coffee break area. \\
        \textbf{User:} when did you see the plants \\
        \textbf{Robot:} According to the observations, plants were seen at different times. However, specific timestamps were not provided for all of these observations. The time a plant was seen in the ``survive" area was at 08:49, one in corridor 2 doesn't have a time, and the one in the reception area wasn't time-stamped either. \\
        \textbf{User:} did we see people after we left the coffee area \\
        \textbf{Robot:} The observations do not provide a sequential order of the events, so it's unclear if we saw people after leaving the coffee break area. \\
      \bottomrule
    \caption{Example using the GPT-triples model. This dialogue was given by the participant a score of 3 for both statements. This dialogue illustrates questions to which the triples-based model had difficulty generating responses, namely concerning the chronology of perceived objects and explored locations.}
    \label{table:example_dialogue1}
\end{tabular}
\end{figure}





\begin{figure}[p]
\begin{tabular}{p{0.99\linewidth}} \toprule
      \textbf{GPT-verbal} \\ \hline
        \textbf{User:} what's up \\
        \textbf{Robot:} As a robot, I don't have personal experiences, but I have recently completed the exploration of an office floor. \\
        \textbf{User:} \textit{[Utterance not recognized]} \\
        \textbf{Robot:} I'm sorry, but your previous message was not understood. Could you please provide more details or rephrase the query? \\
        \textbf{User:} can you mention all the rooms we went to \\
        \textbf{Robot:} Yes, we visited the starting area, hallway, print the room, Lisa's office and the break room. \\
        \textbf{User:} how often do we go in the hallway \\
        \textbf{Robot:} We visited the hallway once during the exploration. \\
        \textbf{User:} did you see any plants \\
        \textbf{Robot:} Yes, I observed a plant in the hallway during the exploration. I may have also seen a plant in Lisa's office, but I'm not certain. \\
        \textbf{User:} did you see any fruits \\
        \textbf{Robot:} I may have seen fruit in the hallway and the break room, but the observations are uncertain. \\
        \textbf{User:} \textit{[Utterance not recognized]} \\
        \textbf{Robot:} I'm sorry, but your previous message was not understood. Could you please provide more details or rephrase the query? \\
        \textbf{User:} did you see any apples \\
        \textbf{Robot:} The observations do not provide information on whether I saw any apples during the exploration. \\
        \textbf{User:} did you see any paintings \\
        \textbf{Robot:} I may have seen a painting in the hallway, but the observation is uncertain. \\
        \textbf{User:} can you dance for me \\
        \textbf{Robot:} As a text-based AI, I'm afraid I don't have the capability to dance. \\
        \textbf{User:} did you see any paintings in leases office \\
        \textbf{Robot:} The observations do not provide information on whether I saw any paintings in Lisa's office during the exploration. \\
        \textbf{User:} did we go to lashes office \\
        \textbf{Robot:} There is no mention of a place called ``lashes office" in the observations. We did go to a place called ``Lisa's office". \\
        \textbf{User:} did you see a ping pong table \\
        \textbf{Robot:} The observations do not provide information on whether I saw a ping pong table during the exploration. \\
        \textbf{User:} did you see a coffee machine \\
        \textbf{Robot:} The observations do not provide information on whether I saw a coffee machine during the exploration. \\
      \bottomrule
          \caption{Example using the GPT-verbal model. This dialogue was given a score of 4 for statement 1 and 3 for statement 2. this dialogue featured a number of speech recognition errors. Additionally, the model responded to several questions with an expression of uncertainty about the observations. Notably, the final turn was accurate in regard to the observations as no coffee machine was recorded by the object detection model.}
    \label{table:example_dialogue2}
\end{tabular}
\end{figure}
\renewcommand{\figurename}{Figure}

\section{Discussion}
\label{sec:discussion}



\subsection{Assessment of factuality}

Variability in how participants approached the tour and what was shown to the robot also affected the results. With regard to the first question, participants appeared to differ in their criteria of whether a statement by the robot was ``correct" or not. A score of ``5" being ``Always" factual could be interpreted to require that the model hallucinated precisely zero times, while other participants interpreted the statement as that the robot's responses as a whole were always factual. 

One alternative evaluation strategy is to evaluate the factuality of the responses against a ground truth. However, defining such a ground truth is particularly difficult in human--robot interaction, as the agent must rely on imperfect and often error-prone sensors to perceive its physical environment. 


\subsection{Question types}

In general, the participants tended to focus quite heavily on asking questions related to what was seen in the tour, as opposed to distances traveled or relative locations. This tendency is natural given that it is simpler to evaluate factuality with respect to seen objects as opposed to less easily memorable aspects such as number of robot turns or exact times rooms were entered. This kind of question was handled well by all models. The questions posed by the users can be divided into several categories:

\begin{itemize}
    \item Polar questions (e.g. ``Did you see a chair in the office?", ``Did we go to a room called office?")
    \item Wh-questions (``What did you see in the office?", ``What was the second room we visited?")
    \item Indirect / open-ended questions (``How about after that?", ``Please summarize the tour.")
\end{itemize}

Moreover, these questions principally centered on \textit{1) Visual recognition} (as noted above) and \textit{2) Routing and path description}, along with \textit{3) Requests for summary}, which came exclusively at the start of dialogues. Polar questions requiring a yes or no answer were typically handled well by all models (although as noted above the LLaMA model encountered occasional repeated failures within a dialogue). The advantages of our proposed model are best illustrated in questions which centered on the second category of questions, thereby taking into account temporal ordering of locations or probability of objects. Answers to indirect questions were often improved in this scenario, as the semantic triples model was unable to make use of timestamps to discern the order of events. This element of information was also important when the model was asked about distances travelled, given that the information was split between locations. 

\subsection{Comparison between approaches}

A general benefit of verbalization compared to semantic triples is that the model has access to natural language describing the observations that it can directly use in its answers. This is especially evident in the GPT verbal model’s ability to express the likelihood of having seen objects. While the structured representation presented the same data to the model (and in a more precise form with respect to probabilities), the LLM was not able to effectively use this aspect of the data. Thus, verbalization avoids relying on the model to make logical inferences beyond its ordinary capabilities. 

With regards to temporal data, a verbalized form of the information avoids the necessity of the model estimating time spans across disjoint facts from the graph. Related to this, when the probability of the observation is presented to the LLM in natural language, the model is not required to translate that uncertainty to a specific linguistic expression. This translation does come with some design judgment: Specifically, what is considered to be certain, likely, unlikely, or impossible? These judgments do however allow the verbalization to be sensitive to what a human would consider reasonable estimations of likelihood, rather than relying on an a priori estimation by the LLM. 

Another benefit of our approach is that the relation of the underlying information to the model prompt is explicit and human comprehensible. Common sense intuitions which are not easily understood by LLMs within structured information can be made explicit in the process of verbalization. The advantage of the models’ strong fluidity in generating responses is preserved while explicitly representing factual information in the state graph and thereby the verbalized graph document. Errors arising from failures in the underlying modules are explicit and more easily uncovered compared to dense vector representations which require specialized methods to decipher.

\subsection{Non-linguistic knowledge}

The reliance of the proposed approach on verbalized knowledge can make the model oblivious to content that is not easily expressed in a sentence. Paralinguistic signals \citep{schuller2013paralinguistics} such as body language, facial expressions or intonation are all crucial elements of communication which are not readily represented in text. Other information like physical feedback in a robotic agent's components may also not be readily represented. Lacking verbalization, integrating these information sources with LLMs will likely remain challenging in the future.

\subsection{Limitations of the experimental study}


Regarding the scope of this study, some limitations must be observed. First, the diversity of the study participants was limited to local researchers and students. Another limitation with respect to experiment participants is that most participants were not native English speakers. Additionally, many local place names were Norwegian and thus difficult for an English ASR system to accurately recognize. Both factors led to various transcription errors that likely affected perceptions of model performance. Further study in different populations and specifically with native English speakers would be a warranted follow-up to our experiments to evaluate the approach in broader audiences.

A further limitation of the experimental setup is the comparative simplicity of the VQA model. The visual module did not distinguish between distinct entities of a single type in a labeled area, merely whether one was present. This setup avoided complex failures arising from difficulties of segmenting objects such as chairs, yet participants were often interested in the number of one kind of entity in an area. Moreover, the VQA model only distinguished entities defined from the outset. While this allowed for a more well-defined set of entities of interest in an office environment, it limited the more creative or unusual possibilities for objects participants might ask about from the tour. Some questions from the participants focused on the number of a particular item seen in the tour despite instruction that this information was not available to the robot. 

Lastly, a more sophisticated VQA model may have been able to determine spatial relationships between entities themselves in a given room. Doing so would require estimation of the robot's field of view and position within the room. When an object was observed at the transition point between two rooms, before and after the user provided a label, any objects in the robot's field of view could be attributed to both locations. In several instances this led to an apparent error of the model stating an object as having been observed in a room when it was seen at the end of its path in that area. 

Our study did not incorporate verbal commands to the robot for movement. This choice was made to facilitate the execution of the office tour. Additionally, the robot also suffers from substantial odometry drift, transformations of spatial data such as plotting the robot path in a building map were not possible. Future work might use more precise measurements of the robot's position and integrate verbal commands for movement to the robot. Lastly, continuous input of audio data to the model instead of relying on an external cue to begin speech recognition is possible.  



\section{Conclusion}
\label{sec:conclusion}
In this work, we have presented a model of response generation in human-robot interaction where situated, dialogue-specific data is represented with a graph. As discussed in the preliminaries, graphs present a number of beneficial attributes that are useful for representing these kinds of information. Namely, dialogue systems benefit from the ability to represent diverse types of information in the same structure along with semantic relationships between them. This structure also allows for flexible and effective transformation to natural language that expresses common sense intuitions about structured data. Importantly for embodied agents, a dynamic knowledge graph can be continuously updated in response to real-time information. Using these attributes, we presented a model which represents a dialogue state of diverse and continuously changing information about the physical world as graph of visual, spatial, and natural language data with temporal and probabilistic attributes. The information in this graph is converted to natural language descriptions for use by large-language models. To do so, a simple parameterized function converts paths within the graph to sentences describing the semantics of the relations described by the path. This function is optimized in a few-shot learning approach on manually annotated data.

To assess the benefit of this approach, we have also presented a study in human-robot interaction where the robot responses are generated with an LLM based on a prompt that includes a verbalized representation of the  graph. This study consisted of users taking a robot for a tour of an office environment and subsequently discussing the tour with the robot. The users then rated the robot's ability to accurately, concisely, and naturally summarize the tour. For each dialogue, the robot made use of two LLMs, namely GPT-4 and LLaMA. For the LLaMA model and one version of GPT-4, the graph information was presented to the LLM as verbalized form of the dialogue state graph. 
While LLMs can make use of diversely formatted text, our study indicates that the GPT-4 model likely has improved ability to respond factually when presented data as a natural language document describing the graph information recorded in the graph. 

While a number of limitations exist within our study, the simplicity of the conversion from graph to natural language allows for the expression of relations between multimodal data in an intuitively useful way for LLM-based response generation, perhaps particularly in HRI settings where in-domain data is scarce and expensive to acquire. 

The impact of differing graph verbalization strategies also likely warrants further study, as graph structures and attributes can be verbalized in diverse ways. A promising area for future work is to determine which translations to surface forms produce the best performance improvements in downstream LLMs. These performance gains may also depend upon the performance of underlying modules such as visual-question answering models and robot sensors. 

Different choices of embodied agents and physical sensors may allow for a more precise and wider variety of information to verbalize. Longer explorations of the environment or more fine-grained sensory information may require the incorporation of notions of \textit{relevance} in order to reduce the amount of information presented to the LLM. Only a small part of the knowledge graph may be necessary to generate a response.  
Retrieval-based generation \citep{lewis2020retrieval,li2022survey,walker2023retrieval} would enable a \textit{contextualized} verbalization of the graph, where only the most relevant parts of the graph are retrieved and included in the input to the LLM response generator.  


\bibliography{references,anthology}

\appendix

\section{VQA Objects}
\label{sec:appendixA}
\textit{"a person", "a woman", "a man", "a computer", "a desk", "a table", "a coffee mug", "a chair",
        "a whiteboard", "a garbage can", "a door", "a window", "a plant", "a fire extinguisher",
        "a pillar", "fruit", "apples", "bananas", "flowers", "magazines", "books", "a computer mouse",
        "a couch", "a TV", "a wastebin", "a pen", "a pencil", "scissors", "folders", "a light switch", "cables",
        "notebooks", "paper", "a printer", "a bookshelf", "bookshelves", "a painting", "a camera", "food", "cake",
        "a light", "a lamp", "a stapler", "a red folder", "a blue folder", "a yellow folder", "a green folder",
        "a broom", "a blue curtain", "a wooden pallet", "a marker", "a copy machine", "cardboard boxes"} \\

\end{document}